\pdfoutput=1

\documentclass[11pt]{article}

\usepackage[final]{acl}

\usepackage{times}
\usepackage{latexsym}
\usepackage{stfloats}

\usepackage[T1]{fontenc}
\usepackage[hang,flushmargin]{footmisc}

\usepackage[ruled, linesnumbered, lined, vlined,ruled]{algorithm2e}

\usepackage[utf8]{inputenc}

\usepackage{microtype}
\usepackage{inconsolata}

\usepackage{graphicx}
\usepackage{subcaption}
\usepackage{float}
\usepackage{color} %
\usepackage{mdframed}
\usepackage{enumitem}
\usepackage{framed}
\usepackage{imakeidx}
\usepackage{siunitx} 

\usepackage{amsmath,amsfonts,bm}

\def\eqref#1{equation~\ref{#1}}

\def\1{\bm{1}}

\DeclareMathAlphabet{\mathsfit}{\encodingdefault}{\sfdefault}{m}{sl}
\SetMathAlphabet{\mathsfit}{bold}{\encodingdefault}{\sfdefault}{bx}{n}

\def\bb{\textcolor{blue}}
\definecolor{ggg}{RGB}{26,179,0}

\usepackage{authblk}
\usepackage{booktabs}
\usepackage{multicol}
\usepackage{multirow}
\usepackage{wasysym}
\usepackage{xcolor}
\definecolor{custom_pink}{RGB}{207,103,195}

\title{Add-One-In: Incremental Sample Selection for Large Language Models via a Choice-Based Greedy Paradigm}

\author{
 \textbf{Zhuo Li\textsuperscript{1,2,3}},
 \textbf{Yuhao Du\textsuperscript{1,2,3}},
 \textbf{Xiaoqi Jiao\textsuperscript{4}},
 \textbf{Yiwen Guo\textsuperscript{5}},
 \textbf{Yuege Feng\textsuperscript{6}},
 \textbf{Xiang Wan\textsuperscript{1,2}},
\\
 \textbf{Anningzhe Gao\textsuperscript{1,2}},
 \textbf{Jinpeng Hu\textsuperscript{7}}\thanks{Corresponding author, jinpenghu@hfut.edu.cn},
 \\
 \textsuperscript{1} Shenzhen International Center for Industrial and Applied Mathematics, \\
 \textsuperscript{2} Shenzhen Research Institute of Big Data, \\
 \textsuperscript{3} The Chinese University of Hong Kong, Shenzhen, \\
 \textsuperscript{4} LIGHTSPEED STUDIOS, 
 \textsuperscript{5} Independent Researcher, \\
 \textsuperscript{6} Birmingham City University,
 \textsuperscript{7} Hefei University of Technology
}

\begin{document}
\maketitle

\begin{abstract}
Selecting high-quality and diverse training samples from extensive datasets plays a crucial role in reducing training overhead and enhancing the performance of Large Language Models (LLMs). However, existing studies fall short in assessing the overall value of selected data, focusing primarily on individual quality, and struggle to strike an effective balance between ensuring diversity and minimizing data point traversals. Therefore, this paper introduces a novel choice-based sample selection framework that shifts the focus from evaluating individual sample quality to comparing the contribution value of different samples when incorporated into the subset. Thanks to the advanced language understanding capabilities of LLMs, we utilize LLMs to evaluate the value of each option during the selection process. Furthermore, we design a greedy sampling process where samples are incrementally added to the subset, thereby improving efficiency by eliminating the need for exhaustive traversal of the entire dataset with the limited budget. Extensive experiments demonstrate that selected data from our method not only surpasses the performance of the full dataset but also achieves competitive results with recent powerful studies, while requiring fewer selections. Moreover, we validate our approach on a larger medical dataset, highlighting its practical applicability in real-world applications. Our code and data are available at \url{https://github.com/BIRlz/comperative_sample_selection}.
\end{abstract}

\section{Introduction}

Large Language Models (LLMs)~\cite{brown2020languagemodelsfewshotlearners, touvron2023llamaopenefficientfoundation, openai2024gpt4technicalreport} have demonstrated exceptional success in AI~\cite{chiang2023vicuna,hu2025agentmentalinteractivemultiagentframework,li-etal-2025-self,dai2025psyche,hu2023simple,hu2022graphenhancedcontrastivelearning,hu2025beyond,hu2024psycollm}. Following the knowledge-based pre-training stage, human-oriented supervised fine-tuning (SFT)~\cite{wei2022finetunedlanguagemodelszeroshot,du2025simplifyrlhfrewardweightedsft} significantly improves LLMs with the most increased performance. However, a huge number of parameters and high complexity of these models also lead to substantial computational and financial demands during SFT, especially when faced with extensive training data.

%
Recently, some studies have indicated that not all instruction data equally contribute to fine-tuning, with a most representative subset often sufficient to match or surpass the performance of LLMs tuned on full datasets \cite{zhou2023limaalignment}. 
%
%
Moreover, many work have shown that the diversity and quality of SFT data are crucial for fully unleashing the potential of LLMs~\cite{ouyang2022traininglanguagemodelsfollow,xu2023wizardlmempoweringlargelanguage}.
%
Therefore, there is a growing interest in developing sample selectors to identify optimal subsets for more efficient SFT. The construction of data selectors relies on the design of selection criteria, considering both the sources of quality labels and approaches to obtain them, which form the fundamental difference for judging data quality. As highlighted by ~\citet{liu2024essencediscarddrossrethinking}, current approaches broadly adopt two strategies. The first one leverages sample internal information and solely relies on its intrinsic characteristics~\citep{li2024quantityqualityboostingllm}. The second strategy incorporates quality labels derived from external sources such as LLM preference judgments~\citep{chen2024alpagasustrainingbetteralpaca,liu2024what} or continuous influence scores quantifying a sample’s impact on model behavior~\citep{xia2024lessselectinginfluentialdata,cao2024instructionmininginstructiondata}.

\begin{figure}[!t]
    \centering
    \includegraphics[width=1\linewidth]{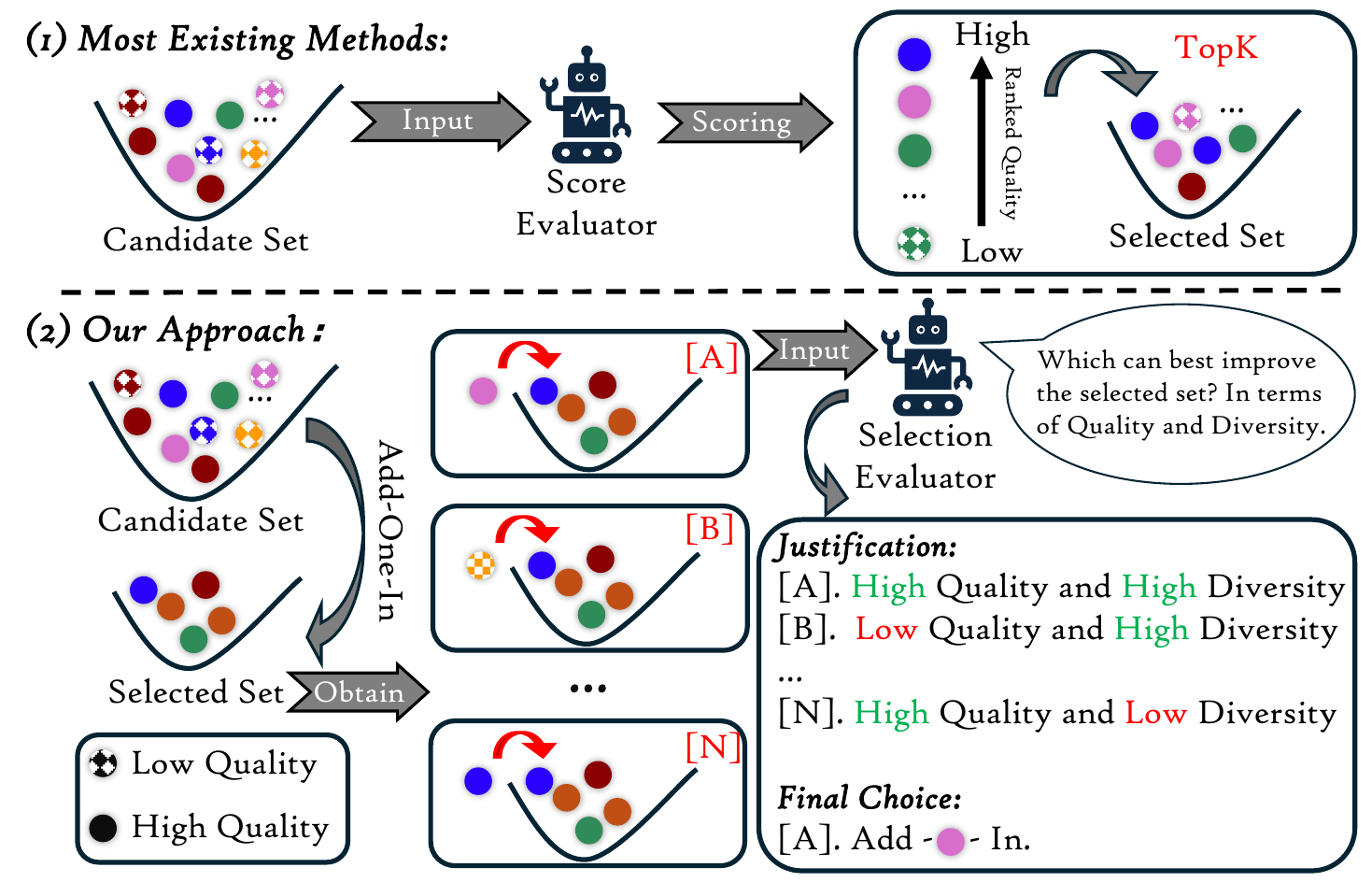}
    \caption{Colors represent data categories, while solid / dapple circles respectively stand for high- / low-quality data. (1) Most existing methods adopt a pointwise approach to produce a subset with top-$K$ representative samples but ignore relationships among them. (2) Our method considers the quality and diversity contribution of each sample to the selected subset. For example, although both \bb{\CIRCLE} and \textcolor{custom_pink}{\CIRCLE} exhibit high-quality in candidate set, incorporating \textcolor{custom_pink}{\CIRCLE} into the selected subset is essential for enhancing its diversity, as the current selected subset already contains \bb{\CIRCLE}.}
    \label{fig_motivation}\vspace{-1em}
\end{figure}

Although these models have achieved considerable improvement, they usually suffer from several issues. Scoring results often lack sufficient differentiation, with many instances receiving identical scores like AlpaGasus~\cite{chen2024alpagasustrainingbetteralpaca} and DEITA~\cite{liu2024what}. On the other hand, \citet{li2024llmsasjudgescomprehensivesurveyllmbased} has highlighted that effective pointwise scoring is inherently more challenging for LLMs than pairwise or listwise evaluation. Moreover, independent scoring for individual sample may overlook the internal relationships within the selected subset, which play a crucial role in ensuring the diversity of the data selection. Although some studies~\cite{liu2024what, liu2024tsdsdataselectiontaskspecific} also propose to improve diversity with the help of explicit diversity measurement like cosine distance and optimal transport~\cite{NIPS2013_af21d0c9}, they typically rely on traversing the entire dataset first, leading to substantial selection demands and time consumption.

Motivated by the Shapley value~\cite{shapley1951notes}, which measures the contribution of each individual in a cooperative setting, we propose a novel choice-based greedy selection framework, as shown in Fig.~\ref{fig_motivation}. Our framework shifts from traditional individual scoring to an option selection strategy. Specifically, we generate multiple candidate options by combining samples from the current selected subset with those from the remaining candidate set. These options are then evaluated and compared to identify the most suitable one. Unlike conventional methods that rely on explicit utility functions for evaluation, we leverage the language understanding capabilities of LLMs through a carefully designed prompt to assess each option's value in terms of both quality and diversity.
To reduce computational overhead, we employ a sampling process to construct these options, where we sample two lists with a fixed window size: one from the selected subset and the other from the remaining dataset, approximating their respective ensembles. We then apply a greedy strategy to iteratively select the optimal option until the desired subset size is achieved.
This approach accounts for the inter-dependencies among samples within the selected subset through an option-based selection mechanism. It also alleviates the need to access and traverse the entire dataset, thereby efficiently identifying a higher-quality and more diverse subset within a limited data budget.

Extensive experiments prove the effectiveness of our method. On the Alpaca instruction tuning dataset~\cite{alpaca}, our method shows that less than 10\% of the data can outperform the model trained on the full dataset, and it also exhibits higher effectiveness compared to SOTA methods. Furthermore, we validate the effectiveness of our approach on a larger medical dataset, showcasing its practical applicability in real applications. 

    
    
    

\section{Related Work}
The field of instruction sample selection for LLMs has seen significant advancements in recent years, driven by the need to improve model performance (i.e., safety~\cite{du2024detectingaiflawstargetdriven}) while reducing training costs~\cite{liu2024essencediscarddrossrethinking}. Early efforts focus on general data selection methods, often relying on human-designed features or simple heuristics to identify high-quality and more diverse samples. For instance, Instruction-Mining~\cite{cao2024instructionmininginstructiondata} and InstructionGPT-4~\cite{wei2023instructiongpt4200instructionparadigmfinetuning} utilize linguistic indicators and GPT-4 scores to guide the selection process, aiming to capture the quality of data through surface-level features. However, these methods often fell short in capturing the nuanced interactions between data and model performance. A more targeted approach emerged with the introduction of model- and data-centric selection criteria, where methods like IFD~\cite{li2024quantityqualityboostingllm} and SuperFiltering~\cite{li2024superfilteringweaktostrongdatafiltering} leverage internal information from the candidate dataset and the target model itself to select data, such as instruction following difficulty and perplexity of each sample. However, these methods are highly model-dependent and rely on the pre-experience training of the target LLM. 

Recent advancements have further refined data selection by incorporating external information and producing quality labels. For example, AlpaGasus~\cite{chen2024alpagasustrainingbetteralpaca} and DEITA~\cite{liu2024what} introduce discrete quality labels derived from LLM preferences, while methods like LESS~\cite{xia2024lessselectinginfluentialdata} use continuous quality labels based on sample influence. However, the formers produce identical quality scores, leading to insufficient differentiation among samples, and the latter can only handle specific-task datasets. Besides, although \citet{liu2024what} and~\citet{liu2024tsdsdataselectiontaskspecific} also encourage diversity during the selection, similar to the methods of generating scores, they still rely on traversing the complete dataset first and require to explicitly design a complex diversity measurement function. Unlike these works, we eliminate pre-computed labels and explicit metrics through LLM-powered dynamic combinatorial evaluation, achieving fine-grained differentiation, task-agnosticity, and more time efficiency without full traversal.

In addition, there are also various selection methods that target on facilitating LLMs' pre-training, like D4~\cite{tirumala2023d4improvingllmpretraining}, DSIR~\cite{xie2023dataselectionlanguagemodels}, and QuRating~\cite{wettig2024quratingselectinghighqualitydata}. Besides, \citet{agrawal2022incontextexamplesselectionmachine} and \citet{nguyen2023incontextexampleselectioninfluences} expand sample selection into in-context learning and machine translation, while we focus on SFT.


\section{Method}

Let $\mathcal{D} = \{ x_1, x_2, \ldots, x_N \}$ denote an instruction tuning dataset with size $N$. Usually each sample $x_i$ is a triplet $\{\text{Instruction}, [\text{Input}], \text{Answer}\}$, where $[\text{Input}]$ is an optional part associated with the instruction. Given a trainable LLM parameterized by $\theta \in \mathbb{R}^d$, we denote $\theta_{\mathcal{D}}$ as the instruction fine-tuned LLM $\theta$ on dataset $\mathcal{D}$. Our objective is to effectively find a subset $\mathcal{A} \subseteq \mathcal{D}$ ($|\mathcal{A}|=K\ll|\mathcal{D}|$) with $K$ data budget, such that each selected sample in $\mathcal{A}$ satisfies specific criteria defined by a selection function $F(\cdot)$. Moreover, $\theta_{\mathcal{A}}$ can achieve comparable performance than $\theta_{\mathcal{D}}$ on various downstream tasks.

\subsection{Warming-Up}
As mentioned above, the diversity of $\mathcal{A}$ and the quality of each sample $x_i \in \mathcal{A}$ affect the performance of the fine-tuned model $\theta_{\mathcal{A}}$. In order to obtain a desirable subset, we begin with the definition of the diversity contribution of a sample $x_i$ to a subset $\mathcal{A}$, where we leverage the following key motivation: \textit{the diversity of a set should depend on how varied or different its elements are.}

Specifically, given an initial set $\mathcal{A}$ and a candidate set $\mathcal{B}$ (initially $\mathcal{B}=\mathcal{D}$) as a start point, for a sample $x_i \in \mathcal{B}$, we define its diversity contribution to the subset $\mathcal{A}$ through a marginal gain $\Delta F(x_i|\mathcal{A})$, where larger $\Delta F(x_i|\mathcal{A})$ indicates that sample $x_i$ can increase the diversity of the set $\mathcal{A}$. We aim at finding the most valuable samples in $\mathcal{B}$ by evaluating $\Delta F(x_i|\mathcal{A})$ for each candidate sample $x_i$ and select the one that maximizes this marginal gain with the help of a carefully designed selection function $F(\cdot)$ by:
\begin{equation}
\begin{aligned}
x^* &= \arg\max_{x_i \in \mathcal{B}}\Delta F(x_i|\mathcal{A}) \\ &= \arg\max_{x_i \in \mathcal{B}} \left[ F(\mathcal{A} \cup \{ x_i \}) - F(\mathcal{A}) \right]. 
\end{aligned}
\label{diverse_obj}
\end{equation}
This selection process continues until the subset $\mathcal{A}$ reaches the desired size $K$ and at each iteration, we add the optimal sample $x^*$ into $\mathcal{A}$ and remove it from $\mathcal{B}$. Besides, a desirable selection function $F(\cdot)$ is also expected to have the ability to judge the quality of the input sample $x_i$. By finding such an effective $F$, we can take the quality of samples into account during the process of finding a diversity subset using Eq.~\ref{diverse_obj}, and ultimately obtain such a desirable subset for LLM tuning. However, it's important to note that finding the exact optimal subset $\mathcal{A}^*$ by Eq.~\ref{diverse_obj} should be NP-hard due to the difficulty of the combinatorial nature of the subset selection task and the computational complexity associated with evaluating all possible subsets~\cite{1982Algorithmic}.
\begin{figure*}[!t]

    \centering
    \includegraphics[width=1\linewidth]{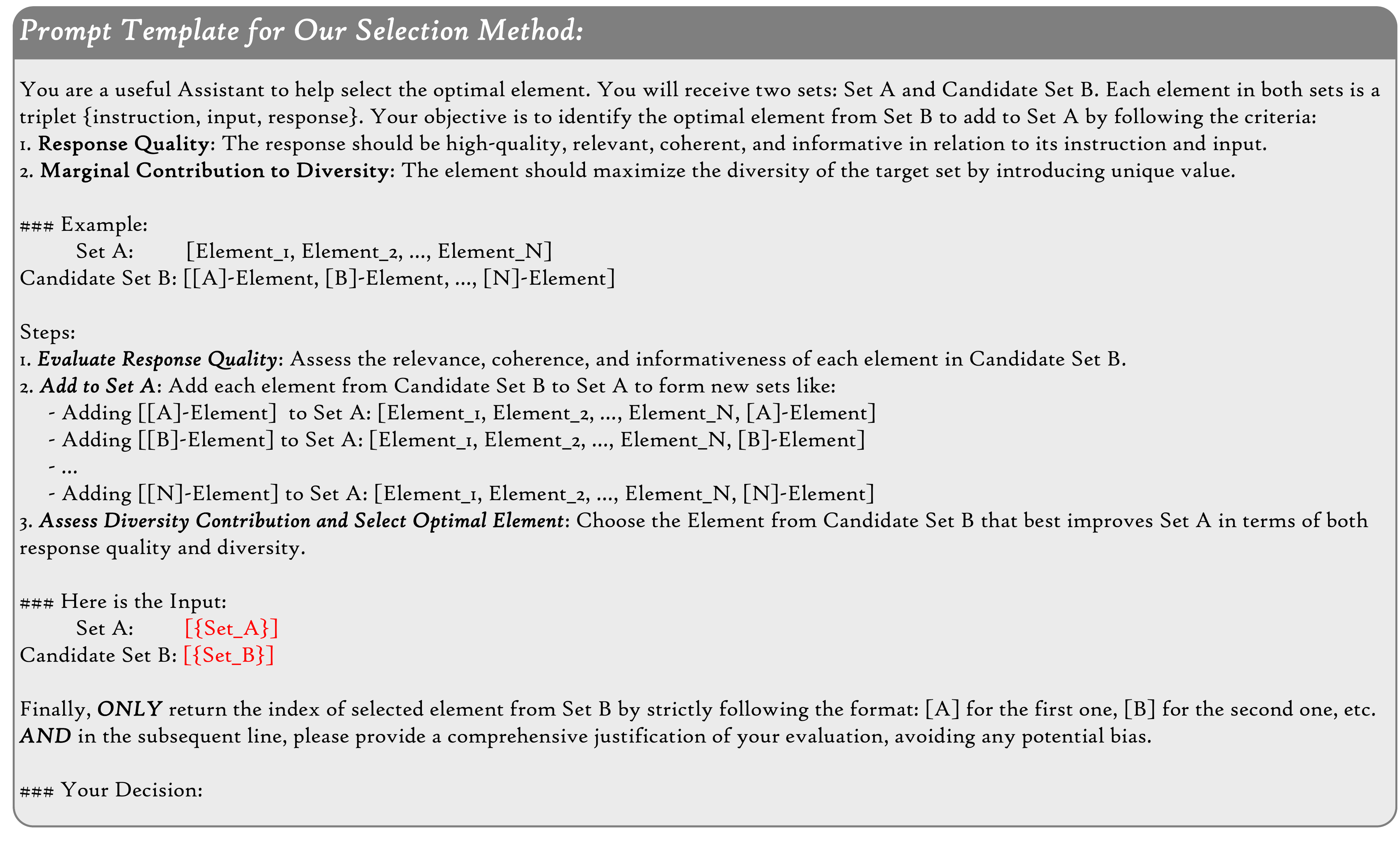}
    \caption{The prompt employed in our method to select the optimal element from the candidate set $\mathcal{B}$ with the help of LLM, considering response quality and diversity contribution to the set $\mathcal{A}$.}
    \label{selection_prompt}\vspace{-1em}
\end{figure*}
\subsection{An LLM-Driven Greedy Method For Contribution-Oriented Subset Selection}\label{our_method}
To efficiently find a desirable subset $\mathcal{A}$ without incurring the computational overhead of evaluating each candidate individually, we propose an LLM-driven greedy method that leverages the expressive capabilities of LLMs in understanding of the input and capable of handling the long-context to select the most valuable sample in a single computation at each iteration. Specifically, in each selection iteration, we firstly fill a selection prompt with $\mathcal{A}$ and $\mathcal{B}$ simultaneously, then query an LLM to make a choice about which element in $\mathcal{B}$ exhibits higher quality and can contribute the most diversity to $\mathcal{A}$. 


As shown in the Fig.~\ref{selection_prompt}, we reformulate the selection obj.~\ref{diverse_obj} by serving an LLM as the implementation of $F$ by such a carefully designed selection prompt. In order to prevent the LLM from failing to understand this complex selection prompt well, resulting in mis-following and ultimately being unable to make effective decisions, we adopt an ICL strategy~\cite{dong2022survey} to assist the LLM in better accomplishing our selection task and finally obtaining an effective optimal element. Finally, we can obtain an optimal element $x^*$ from the candidate set $\mathcal{B}$ by prompting $F_{\text{LLM}}$. 

However, 
given the input length limitation and computational complexity associated with LLMs when processing long contexts, it is impractical to input the complete $\mathcal{A}$ and $\mathcal{B}$ into the LLM all at once.
%
%
Therefore, during each iteration of selection, we randomly select $\mathcal{A}'$ from the current selected subset $\mathcal{A}$ and $\mathcal{B}'$ from candidate set $\mathcal{B}$, in order to avoid limitations and alleviate the traversal overhead. The selection process can be modified as:
\begin{equation}
x^* = F_{\text{LLM}}(\mathcal{A}', \mathcal{B}', \textit{prompt template}),
\label{final_obj}
\end{equation}
where sizes of $\mathcal{A}'$ and $\mathcal{B}'$ are denoted as $L_A$ and $L_B$, respectively. Our LLM-driven greedy selection process are summarized in App.~\ref{alg_selection}, where our method begins with a random initialization of $\mathcal{A}$ and without specific statement, we set $L_A=L_B=20$.

\begin{algorithm}[h]
\SetAlgoNlRelativeSize{-1}
\caption{Algorithm Process of Our Method.}
\label{alg_selection}
\KwIn{$\mathcal{D}$: Full dataset, an LLM and hyper-parameters: \{$K$: Desired subset size, $L_A$ and $L_B$: window size.\}}
\KwOut{$\mathcal{A}$: Selected subset}
\SetKwFunction{FMain}{GreedySampling}
\SetKwProg{Fn}{Function}{:}{}
    {\textbf{Initialize} subset $\mathcal{A} \gets \text{RandomSample}(\mathcal{D}, L_A)$}\;
    {\textbf{Initialize} candidate $\mathcal{B} \gets \mathcal{D}\setminus\mathcal{A}$}\;

    \While{$|\mathcal{A}| < K$}{
        {\textbf{Initialize} local $B' \gets \text{RandomSample}(\mathcal{B}, L_B)$ and $A' \gets \text{RandomSample}(\mathcal{A}, L_A)$}\;

        {\textbf{Obtain} the optimal sample index by $x^* = F_{\text{LLM}}(\mathcal{A}', \mathcal{B}', \textit{prompt template})$}\;
        {\textbf{update} $\mathcal{A} \gets \mathcal{A} \cup \{x^*\}$}\;
        {\textbf{update} $\mathcal{B} \gets \mathcal{B} \setminus \{x^*\}$}\;
    }
    \Return{$\mathcal{A}$}
\end{algorithm}

\subsection{Discussion}\label{sec:discussion}
Our greedy selection method relies on the LLM's ability to comprehend and accurately follow the provided selection instructions, where prior work has shown that LLMs are proficient in understanding and executing detailed prompts~\cite{ouyang2022traininglanguagemodelsfollow, wei2023chainofthoughtpromptingelicitsreasoning}. Besides, our method can be likened to listwise evaluation approaches, where prior research~\cite{Zhuang_2024, hou2024largelanguagemodelszeroshot} have demonstrated that LLMs are capable of performing listwise selection and ranking tasks effectively. These studies have shown that LLMs can assess and rank a list of items based on certain criteria, which aligns with our approach of selecting items that maximize a set function considering both quality and diversity.

Moreover, our method parallels the classical greedy algorithm used in submodular optimization. when an LLM can correctly follow the selection prompt, our greedy algorithm could achieve at least a $(1-{e}^{-\gamma})$ approximation of the optimal solution~\citep{chen2018weakly}, where the parameter $\gamma\in [0,1)$ is a submodularity ratio related to the LLM. This means that our method not only efficiently builds a diverse and high-quality subset but also provides theoretical assurance on the solution's near-optimality. In terms of computational cost, each iteration involves querying the LLM to select the next sample, resulting in a total time complexity of $\mathcal{O}(K\cdot T_{\text{LLM}})$, where $K$ is the desired subset size and $T_{\text{LLM}}$ is the inference time of the LLM. This linear complexity with respect to $K$ makes our method scalable to large datasets and requires fewer accesses to the entire dataset.

\section{Experiment}
\subsection{Instruction Fine-tuning Dataset}
\subsubsection{Settings}
\paragraph{Datasets} We mainly select an effective training subset from the Alpaca dataset that encompasses 52K instruction-following samples~\cite{alpaca}. In order to achieve less biased assessment, we evaluate the performance of our method on four popular testsets: Vicuna~\cite{chiang2023vicuna}, Koala~\cite{zhang2024koalaenhancingspeculativedecoding}, WizardLM~\cite{xu2023wizardlmempoweringlargelanguage} and Self-instruct~\cite{wang2022self}, which totally contains 730 human generated instructions from different tasks and sources to ensure the comprehensive convergence of task types.
\begin{figure*}[!ht]
    \centering
    \includegraphics[width=1\linewidth]{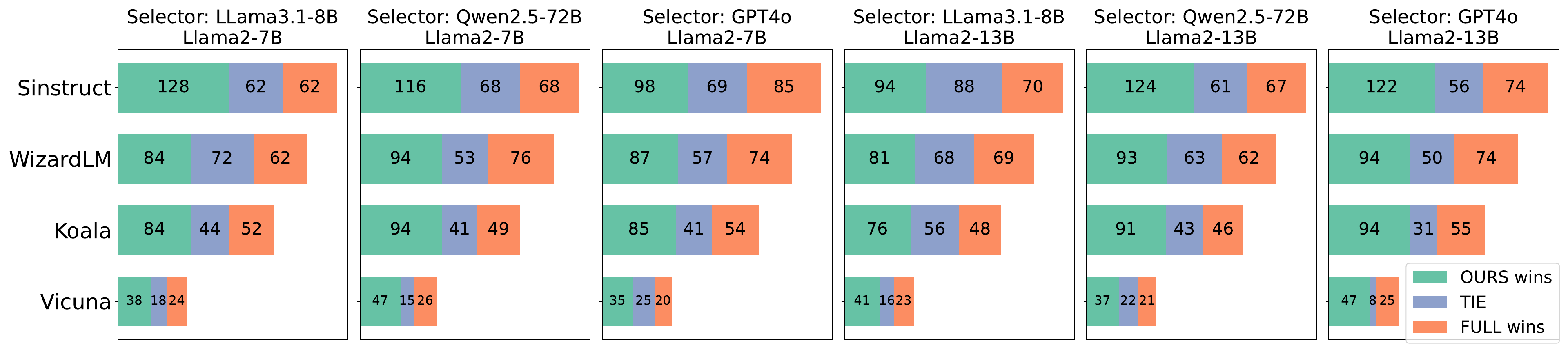}
    \caption{Comparing models fine-tuned on our method (9K) and full data (52K) on Llama2-7B and Llama2-13B with different LLMs as selector.}
    \label{fig:main_exp}\vspace{-1em}
\end{figure*}
\paragraph{Selector} In order to comprehensively evaluate the flexibility of our method, we employ three types of LLM as the selector: 1) Small-size: Llama3.1-8B-Instruct~\cite{touvron2023llamaopenefficientfoundation}; 2) Medium-size: Qwen2.5-72B-Instruct~\cite{yang2024qwen2technicalreport}; 3) Closed-source: GPT3.5 / GPT4o~\cite{openai2024gpt4technicalreport}.

\paragraph{Baseline} We compare the effectiveness of our method with the following methods: 1) Full data; 2) Random; 3) Two popular SOTA methods - AlpaGasus~\cite{chen2024alpagasustrainingbetteralpaca} and IFD~\cite{li2024quantityqualityboostingllm}, both of which require to firstly traverse the entire dataset, then rank, and finally select the top-$K$ highest samples. We mainly consider these two SOTA methods due to the similar scope and experimental settings. For more concise expression, we use the names of these methods to represent the models that are fine-tuned on the corresponding datasets. For instance, ``Full'' is used to denote $\theta_{\text{Full}}$. 
\paragraph{Implementation Details} Refer to App.~\ref{app:implement}.

\subsubsection{Evaluation Metrics} 
\paragraph{Pairwise Comparison} Following the common practice of LLM-as-a-judge, we utilize GPT4o and adopt the evaluation template from MT-Bench~\cite{zheng2023judgingllmasajudgemtbenchchatbot}. 
To alleviate potential positional bias, we present the responses of two models to the judge in two different orders and compare their scores. A model is considered to win only if it does not lose in both orderings. Specifically, we define the outcomes as follows: \textbf{Wins}: Outperforms in both orderings or wins in one and ties in the other. \textbf{Tie}: Ties in both orderings or wins in one and loses in the other. \textbf{Loses}: Lags in both orderings or ties in one and loses in the other.

\paragraph{Benchmark Evaluation} To fully understand how our selected samples influence the fine-tuning when compared with baselines, we also evaluate performance on several popular benchmarks, i.e., MMLU~\cite{hendrycks2021measuringmassivemultitasklanguage}, BBH~\cite{suzgun2022challengingbigbenchtaskschainofthought} and Hellaswag~\cite{zellers2019hellaswagmachinereallyfinish}.

\subsubsection{Results}
\paragraph{Pairwise Comparison with Full (52K)} By following AlpaGasus, which selects the 9K highest-quality samples scored by GPT3.5, we first demonstrate the effectiveness of our method using the same scale of selected samples. 
As shown in Fig.~\ref{fig:main_exp}, our model trained with 9K samples significantly outperforms the model trained on the full dataset under various settings. 
We employ a diverse range of LLMs as selectors and as models to be trained (Llama2-7B-hf and Llama2-13B-hf). 
These results illustrate that our method effectively identifies valuable and diverse samples for instruction fine-tuning, leveraging the powerful language understanding capability of LLMs, regardless of the specific size or family of models used for selection or training.

\begin{figure}[t]
    \centering
    \includegraphics[width=1\linewidth]{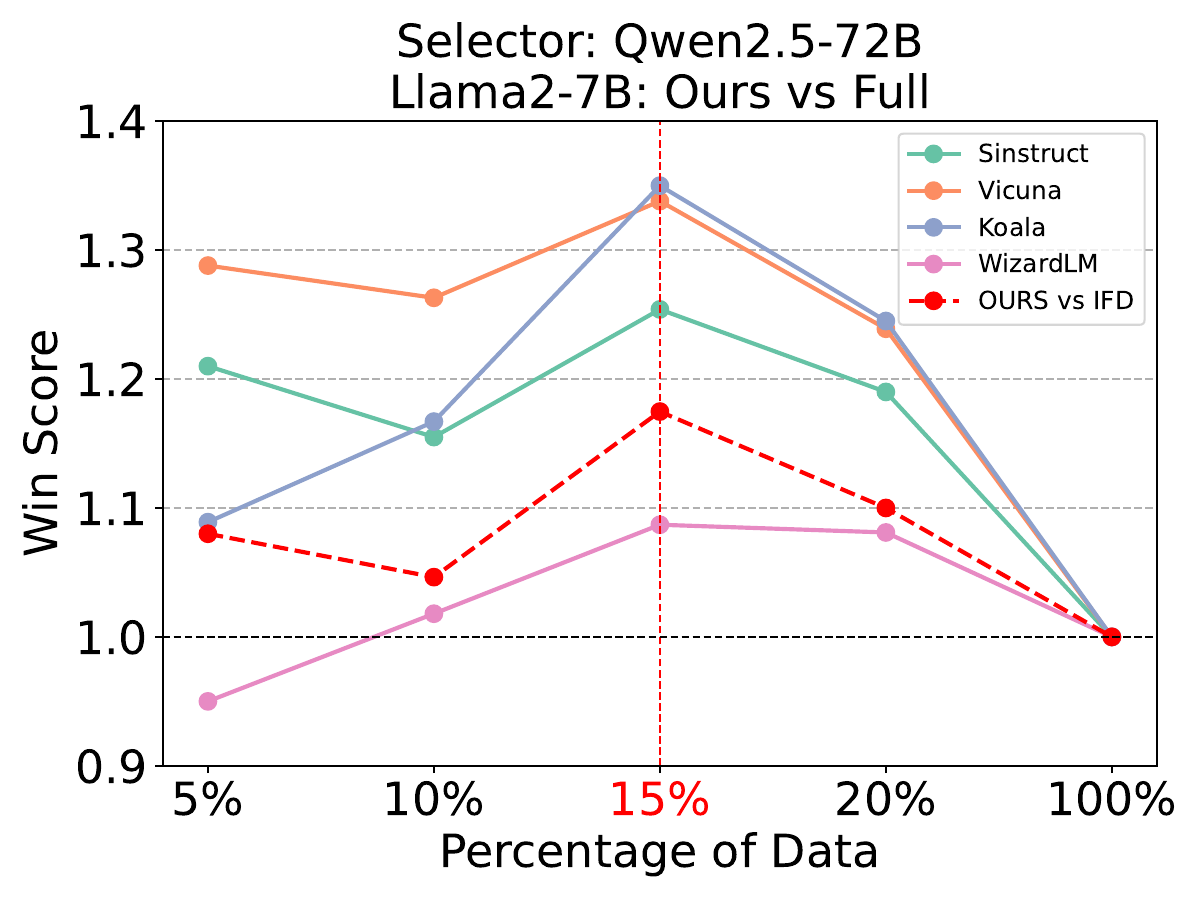}
    \caption{The win score changes with the increasing of data scale by comparing ours with the Full and IFD.}
    \label{fig:compare_IFD}\vspace{-1em}
\end{figure}

Furthermore, we conduct experiments by selecting subsets corresponding to different percentages of the training dataset, ranging from 5\% to 20\%, and compare the average win score changes relative to the full dataset as the amount of data increases. 
%
WS can be calculated as $\frac{\#\text{Win} - \#\text{Lose}}{\#\text{All}} + 1$, where a score greater than 1.0 indicates that the model outperforms the one that is compared against.
As shown in Fig.~\ref{fig:compare_IFD}, with just 5\% of the data it selects, our model can outperform those trained on the full dataset on three of the four test datasets.
As the amount of data increases, our method consistently exceeds the performance of the Full across all four datasets. 
In particular, with 15\% data usage, our model achieves the best performance, with an average win rate of 1.257. 
These results demonstrate that appropriately selecting a subset of the full data is sufficient to enhance the LLM's instruction-following ability, and our incremental method effectively identifies a high-quality and more diverse subset for improved LLM tuning.

\begin{figure}[t]
    \centering
    \includegraphics[width=1\linewidth]{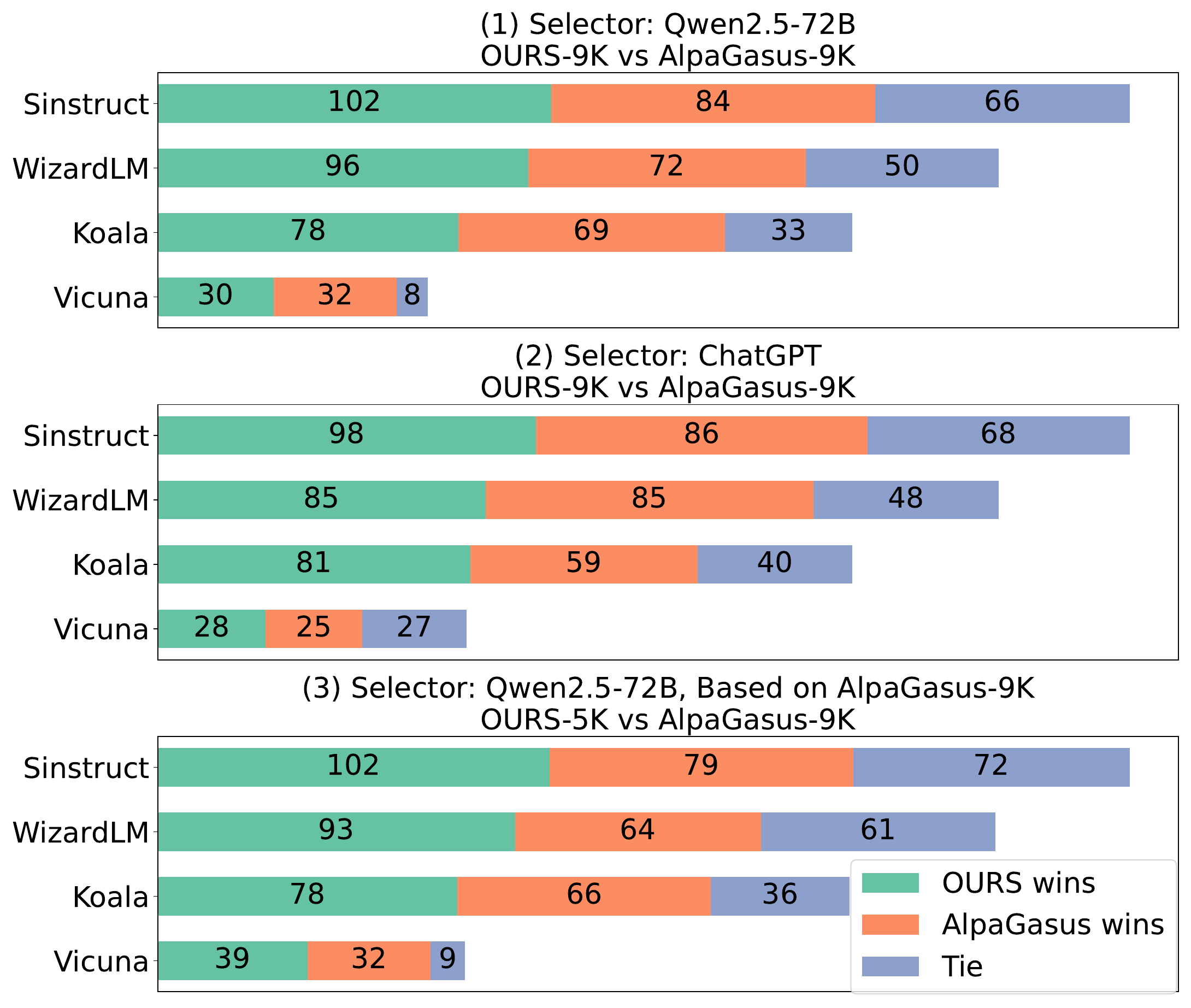}
    \caption{Comparing our method with AlpaGasus under 9K data on Llama2-7B.}
    \label{fig:compare_alpagasus}\vspace{-1em}
\end{figure}

\paragraph{Pairwise Comparison with SOTA methods} Beyond our comparison with the full data, we also evaluate our performance against the SOTA methods - AlpaGasus and IFD. For comparison with AlpaGasus, we select 9K samples for a fair evaluation at the same data scale, due to the nature of AlpaGasus's discrete quality labels. As shown in (1) and (2) of Fig.~\ref{fig:compare_alpagasus}, ours significantly outperforms AlpaGasus across four datasets, indicating our method's potential to select effective training samples without the need for pointwise traversal of the entire dataset. 
Furthermore, when we use the 9K high-quality samples selected by AlpaGasus as the candidate pool and apply our method for further selection, we find that the 5K samples chosen by our method can surpass the performance of AlpaGasus, as shown in (3) of Fig.~\ref{fig:compare_alpagasus}. This in-depth exploration reveals that our method does not conflict with pointwise methods; instead, it can effectively enhance existing methods by enabling the selection of more representative samples through a hierarchical and multi-round screening process, thereby facilitating the SFT of more effective LLMs.

\begin{figure}[h]
    \centering
    \includegraphics[width=1\linewidth]{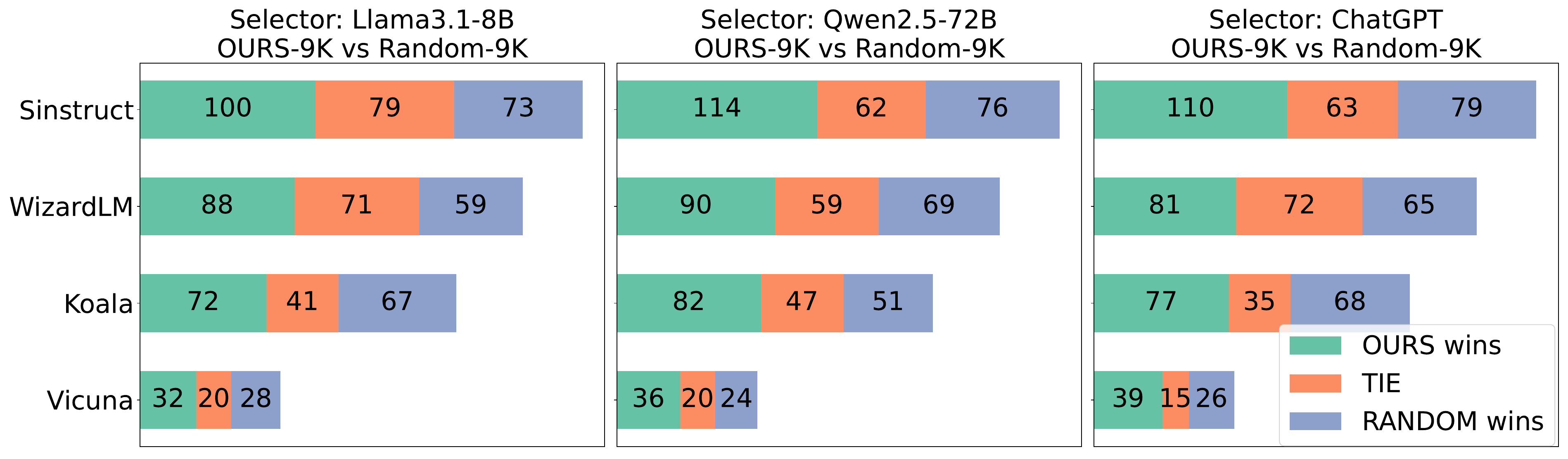}
    \caption{Comparing our method with the Random Baseline with the 9K data on Llama2-7B.}
    \label{fig:compare_random}
\end{figure}

In our comparison with IFD, our method demonstrates competitive performance, as illustrated by the \textit{red dotted} line in Fig.\ref{fig:compare_IFD}. We consistently outperform IFD across various percentages ranging from 5\% to 20\%, achieving an optimal win score of 1.17 at 15\%, which clearly highlights the strength and practical utility of our selection approach. Using the powerful understanding capabilities of LLMs, we can incrementally identify a diverse subset without having to examine the entire dataset at once, thereby greatly improving efficiency by reducing the number of selection times. Additionally, we provide comparisons with the Random baseline in Fig.\ref{fig:compare_random}, where our method consistently outperforms the Random baseline across various settings by significant margins, demonstrating the effectiveness of the proposed method.

\paragraph{Performance on Benchmarks} Following the practice of evaluating LLMs on benchmarks, we also compare our method with Full, Random, AlpaGasus, and IFD, where we set the sample size to 9K for a fair comparison, except for Full. For MMLU, BBH, and Hellaswag, we adopt 0/5-Shot, 3-Shot, and 0-Shot settings, respectively. Tab.~\ref{results_benchmark} suggests that our method provides a promising approach to diverse data selection by achieving competitive performance to baselines with fewer selection times, optimizing both performance and selection cost. {More detailed performance with varying scales of training data can be found in App.~\ref{sec:varying_benchmark}.}

\begin{table}[t]
\centering
\resizebox{0.49\textwidth}{!}{
\begin{tabular}{c|ccccc}
\toprule
\multirow{2}{*}{Benchmark} & \multicolumn{5}{c}{Llama2-7B-hf} \\
 & Full & Random & Alpagasus & IFD & Ours \\ \midrule\midrule
MMLU-0-Shot & 22.09 & 22.51 & 23.25 & 23.12 & \textbf{23.82} \\
MMLU-5-Shot & 45.45 & 46.38 & 46.74 & 46.10 & \textbf{47.44} \\
BBH & \textbf{32.10} & 31.38 & 31.32 & 31.08 & 30.97 \\
Hellaswag & 69.97 & 70.99 & \textbf{71.07} & 70.55 & \textbf{71.07} \\ \midrule
Average & 42.15 & 42.82 & 43.10 & 42.71 & \textbf{43.33} \\ \midrule\midrule
\multirow{2}{*}{Benchmark} & \multicolumn{5}{c}{Llama2-13B-hf} \\
 & Full & Random & Alpagasus & IFD & Ours \\ \midrule
MMLU-0-Shot & 28.07 & 27.21 & 28.38 & 26.79 & \textbf{29.01} \\
MMLU-5-Shot & 53.53 & 52.34 & \textbf{54.38} & 54.28 & 54.28 \\
BBH & 46.40 & 44.58 & 46.21 & 46.25 & \textbf{47.28} \\
Hellaswag & {80.55} & 81.45 & 81.36 & 81.36 & \textbf{81.73} \\ \midrule
Average & 51.38 & 51.89 & 52.58 & 52.17 & \textbf{53.08} \\  \bottomrule
\end{tabular}
}
\caption{The benchmark results of models fine-tuned on different subsets selected by corresponding methods. {More performance and analysis with varying training samples can be found in Tab.~\ref{tab:app_varying}}.}
\label{results_benchmark}\vspace{-1em}
\end{table}

\subsection{Analysis}
In this section, we conduct an in-depth analysis of our method from the perspective of hyperparameter sensitivity, efficiency analysis, characteristics of the selected samples, and case studies.

\paragraph{Ablation Study} As mentioned in Sec.~\ref{our_method}, in each iteration of selection, we randomly select subsets $\mathcal{A}'$ from the current selected set $\mathcal{A}$ with $L_A$ samples and $\mathcal{B}'$ from the candidate set $\mathcal{B}$ with $L_B$ samples to comply with the LLM input limitation. To empirically examine the impact of $L_A$ and $L_B$, we conduct a detailed ablation study by varying the length of set $\mathcal{A}'$ and $\mathcal{B}'$ independently, observing its effect on the diversity and quality of the selected subsets. Additionally, we explore a multi-round selection strategy: initially selecting a larger subset, and then performing further selection on this subset in subsequent rounds until we obtain our desired subset size. This strategy aims to enhance diversity by iteratively refining the subset.

\begin{figure*}[t]
    \centering
    \includegraphics[width=1\linewidth]{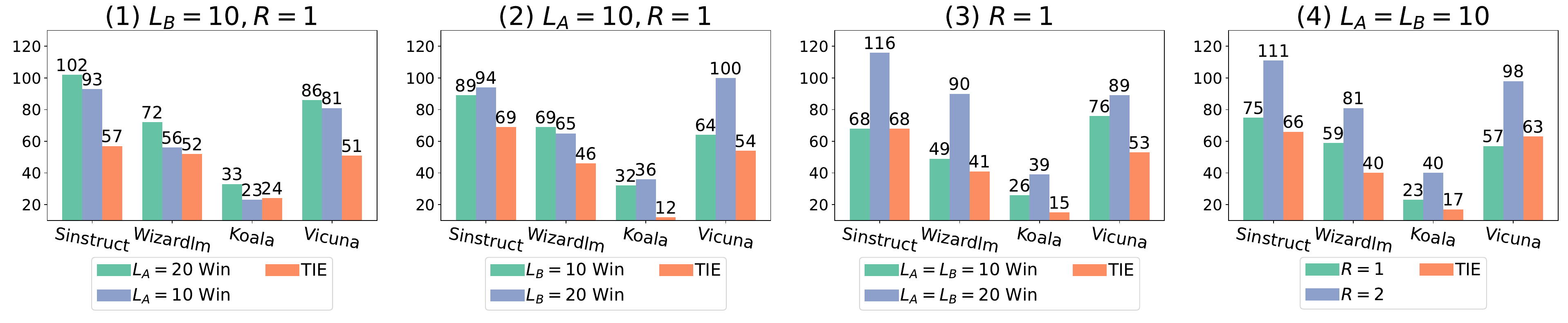}
    \caption{Ablation study on the influence of the lengths of sets $\mathcal{A}'$ and $\mathcal{B}'$ (1, 2, 3), and the number of selection rounds (4). We adopt a pairwise comparison and employ Qwen2.5-72B-Instruct as the selector.}
    \label{results_ablation}\vspace{-1em}
\end{figure*}

Fig.~\ref{results_ablation} compares different configurations in terms of their ability to select better samples and report pairwise comparison results, indicating superiority in diversity and quality of the selected subsets. Our key findings are as follows:
\textit{1) As shown in Fig.~\ref{results_ablation}(1), (2), and (3), increased $L_A$ or $L_B$ leads to better performance consistently across different datasets and model selectors.} 
A larger $L$ provides the LLMs with more context and a broader range of candidates to choose from, which in turn enhances the diversity and quality of the selected samples.
\textit{2) Fig.~\ref{results_ablation}(4) illustrates that our method effectively integrates with a multi-round selection strategy.}
%
%
By adopting a hierarchical, coarse-to-fine approach, we iteratively refine the subset through multiple selection rounds, gradually narrowing the focus to increasingly promising candidates.
The data selected in each round serves as the foundation for constructing new options in the subsequent round, enabling a step-by-step process to identify higher-quality data over iterations.
%
Additional results based on GPT3.5-as-the-selector are in App.~\ref{sec:ablation_study}.

\paragraph{Efficiency Comparison} We investigate whether our method achieves an acceptable trade-off between improved performance and the computational costs associated with the selection, from the perspective of: 1) Time Complexity; 2) Number of Iterations; 3) Practical Time Consumption. Using the Alpaca Dataset (52K), we present an efficiency comparison when selecting the top 10\% of the data. The results in Tab.~\ref{time_consuming} demonstrate that our method significantly reduces selection costs compared to IFD and AlpaGasus. Specifically, operating with a time complexity proportional to $K$ rather than $N$, our method requires significantly fewer iterations and less practical time, due to the greedy selection process that incorporates the powerful capability of LLMs. This not only accelerates the selection procedure, but also maintains high performance, achieving better efficiency without compromising the quality of the selected data. In summary, our method is only proportional to the desired subset size $ K $, embodying the principle of ``selecting only what is necessary''. This characteristic will ensure that our method can guarantee higher efficiency and lower selection costs when faced with extremely large-scale datasets. We provide implementation details in App.~\ref{app:in-deep_analysis}.

\begin{table}[t]
\centering
\resizebox{0.4\textwidth}{!}{
\begin{tabular}{c|ccc}
\toprule
\textbf{Method} & \textbf{TC} & \textbf{NoI} & \textbf{PTC (Minutes)} \\ \midrule\midrule
AlpaGasus & \multirow{2}{*}{$\mathcal{O}(N \times T_{\text{LLM}})$} & \multirow{2}{*}{52,002} & 2466.80 \\
IFD &  &  & 176.80 \\ \midrule
OURS & $\mathcal{O}(K \times T_{\text{LLM}})$ & \textbf{5,200} & \textbf{118.83} \\ \bottomrule
\end{tabular}
}
\caption{Efficiency comparison of different methods on the Alpaca Dataset when selecting the top 10\% data. $T_{\text{LLM}} $ is the time for a single LLM inference. TC, NoI, and PTC indicate the time complexity, number of iterations, and practical time consumption, respectively.}
\label{time_consuming}\vspace{-1em}
\end{table}

\begin{table}[!t]
\centering
\resizebox{0.5\textwidth}{!}{
\begin{tabular}{c|ccc|cc|c}
\toprule
\multirow{2}{*}{Method} & \multicolumn{3}{c|}{Diversity}  & \multicolumn{2}{c|}{Quality} & \multirow{2}{*}{\#Tokens} \\
& TTR $\uparrow$ & MTLD $\uparrow$  & SDI $\downarrow$ & DEITA $\uparrow$ & Helpful $\uparrow$  &  \\ \midrule \midrule
Full (52K) & 95.47 & 7.9352  & 0.1067 & 2.765 & 1.314 & 11.35\\
Random & 95.46 & 7.9421  & 0.1068 & 2.764 & 1.397 & 11.33\\
AlpaGasus & 96.07 & 8.0508 & 0.1086& 2.969 & 2.025 & 10.93 \\
IFD & 96.05 & 8.0444  & 0.1091 &  3.127 & 2.456 & 10.80\\
OURS & \textbf{96.24} & \textbf{8.4449} & \textbf{0.1035} & \textbf{3.210} & \textbf{2.703} & 11.29 \\ \bottomrule
\end{tabular}
}
\caption{Performance of different methods on various text diversity and quality metrics. The OURS method shows the best performance across multiple metrics.}
\label{tab:data_char}\vspace{-1em}
\end{table}

\paragraph{Data Characteristics} We evaluate whether our method identifies a representative subset with higher quality and greater diversity. Tab.~\ref{tab:data_char} presents a comparison using various diversity metrics, such as Type-Token Ratio (TTR)~\cite{article}, Measure of Textual Lexical Diversity (MTLD)~\cite{mccarthy2010mtld}, and Simpson Diversity Index (SDI)~\cite{1997Measurement}, along with quality metrics including DEITA quality score~\cite{liu2024what} and helpful reward value~\cite{dong2023raftrewardrankedfinetuning}. Higher values generally indicate better diversity or quality, except for the SDI. We also report instruction token counts to assess structural similarity to the full dataset. Baseline methods are evaluated on 9K samples versus 52K with average scores reported. We provide implementation details and explanations to these metrics in App.~\ref{sec:characteristics_metrics}.

All three sample selection methods outperform the Full and Random baselines across most metrics, indicating that the full dataset contains redundancies and that strategic sample selection efficiently captures the most informative elements. Compared to AlpaGasus and IFD, our method significantly improves diversity and quality by emphasizing both aspects. Higher TTR and MTLD values in our subset reflect richer vocabulary and lexical structure, while a lower SDI suggests a more richer and balanced dataset. Increased DEITA quality scores and reward values confirm our improvements in both diversity and quality. Additionally, our subset's average token count is similar to that of the full dataset, implying that we maintain the original data's structural characteristics without favoring longer/shorter samples and demonstrating that our method effectively captures the semantic richness of the data. In conclusion, our method selects subsets that are both diverse and representative of the full dataset, offering promising implications for more effective LLM tuning.

\paragraph{Case Study} We highlight three selection cases in App.~\ref{case_ana}.

\subsection{Scaling-Up with Larger Datasets}
We conduct experiments in the medical domain to demonstrate the practical effectiveness of our method. Specifically, based on the HuatuoGPT-sft-data-v1 dataset for SFT samples, we evaluate performance on the Chinese Medical Benchmark (CMB)~\cite{wang2024cmbcomprehensivemedicalbenchmark}, utilizing the Baichuan2-7B-Chat model~\cite{baichuan2023baichuan2}. Given that AlpaGasus requires querying GPT3.5 for sample scoring, we primarily compare our method with the Base model, Full, and IFD. We provide detailed training settings in App.~\ref{sec:medical_exp}.

Tab.~\ref{tab:scale-up} shows that our method consistently improves performance in various data percentages, significantly outperforming both the Full and IFD. For example, with only 20\% of the samples selected by our method, we achieve an improvement of 7.04\% over the Base model and 4.36\% over IFD. These findings highlight potential drawbacks of fine-tuning with the entire dataset and underscore the advantages of our approach in both performance enhancement and time efficiency, suggesting its potential for real-world applications.

\begin{table*}[t]
\centering
\resizebox{0.8\textwidth}{!}{
\begin{tabular}{c|c|cccccc|c}
\toprule
\multirow{2}{*}{Method} & \multirow{2}{*}{\%} & \multicolumn{7}{c}{Chinese Medical Benchmark (\%)}  \\
 &  & Physician & Nurse & Pharmacist & Technician & Disciplines & Exam & Avg. \\ \midrule\midrule
Base & - & 44.20 & 51.69 & 46.06 & 43.83 & 41.56 & 36.56 & 44.29 \\ 
Full & 100 & {46.35} & {57.69} & {50.81} & {43.83} & \textbf{49.81} & {52.00} & {49.38}\\ \midrule
OURS & \multirow{2}{*}{10} & {46.35} & {54.06} & {48.72} & {40.58} & {42.88} & {43.81} & {46.65} \\
IFD &  & {42.55} & {50.19} & {44.75} & \textbf{47.75} & {41.19} & {37.19} & {45.90} \\ \midrule
OURS & \multirow{2}{*}{20} & \textbf{50.65} & \textbf{60.44} & {51.94} & {43.00} & {47.25} & \textbf{52.19} & \textbf{51.33} \\ 
IFD &  & {37.50} & {51.56} & {46.47} & {41.67} & {40.19} & {44.75} & {46.94} \\ \midrule
OURS & \multirow{2}{*}{30} & {49.70} & {58.63} & {52.09} & {41.58} & {46.06} & {50.00} & {50.31} \\
IFD &  & {48.00} & {56.19} & {49.47} & {39.67} & {43.56} & {44.31} & 47.54 \\ \midrule
OURS & \multirow{2}{*}{40} & {49.20} & {60.38} & \textbf{53.66} & {42.58} & \textbf{48.13} & {51.44} & {51.03}  \\
IFD &  & {47.90} & {58.75} & {51.31} & {43.08} & {45.38} & {49.63} & {49.79}  \\ \bottomrule
\end{tabular}
}
\caption{Performance comparison on the Chinese Medical Benchmark (CMB) using different methods and varying percentages of the dataset for fine-tuning. The percentages (\%) indicate the proportion of the dataset utilized. The results are reported across various medical professions.}
\label{tab:scale-up}\vspace{-1em}
\end{table*}

\section{Conclusion}
We propose a novel choice-based sample selection paradigm that shifts the focus from individual scoring to comparing the contribution of each sample when incorporated into a subset. By leveraging the powerful understanding capabilities of LLMs, we are able to simultaneously consider both quality and diversity during the sample selection process. Moreover, we design a greedy process that incrementally builds the subset, which not only eliminates the need to traverse the entire dataset but also significantly reduces selection overhead. Extensive experiments on Alpaca dataset and medical application demonstrate that our method selects more representative subsets with improved selection efficiency compared with SOTA methods, showing as a promising direction for efficient sample selection.

\section{Acknowledgments}
This work was supported by the Guangxi Key R\&D Project  (No. AB24010167), the Project (No. 20232ABC03A25), and the Futian Healthcare Research Project (No.FTWS002). This work was also supported by the National Natural Science Foundation of China (NSFC) under Grant 62402158.

\section{Limitations and Future Work}
While our proposed choice-based sample selection paradigm demonstrates promising results, it is important to acknowledge several limitations and potential areas for future improvement. A significant limitation of our approach is its reliance on LLMs whose capabilities and biases directly impact the sample selection process. The rationality and effectiveness of selecting samples are contingent upon the LLM's ability to accurately assess and compare data points. However, LLMs may have inherent biases inherited from their training data, which can inadvertently influence the selection process~\cite{wang2023largelanguagemodelsfair,ko-etal-2020-look}. Another limitation lies in the need to manually adjust hyperparameters such as the sizes of sets $\mathcal{A}'$ and $\mathcal{B}'$, as well as the design of prompts used to elicit evaluations from the LLM. Selecting the sizes of $\mathcal{A}'$ and $\mathcal{B}'$ impacts the granularity and breadth of the selection process, while the choice of prompts influences the quality of the LLM's assessments.

In the future, enhancing the greedy sampling process by incorporating more sophisticated heuristic methods instead of random selection offers a promising direction. Besides, how to extend my approach to online and sequential data scenarios remains an open challenge. For example,  developing adaptive sampling strategies would significantly enhance the practicality of our method. This could involve designing algorithms capable of making immediate selection decisions in real-time applications, ensuring sustained model performance in environments where data arrives continuously.

\bibliography{acl_latex}

\clearpage
\onecolumn
\appendix

\section{Experiment}\label{sec:appendix}

\subsection{Implementation Details and Cost}\label{app:implement}
\paragraph{Implementation Details} We mainly evaluate the performance of the selection methods on Llama2-7B-hf and Llama2-13B-hf, where we adopt the same training configuration as the original Alpaca using the Stanford codebase\footnote{\small\url{https://github.com/tatsu-lab/stanford_alpaca/tree/main}}. During inference, we employ vLLM~\cite{kwon2023efficient} to help speed up the generation, where we set the sampling temperature = 0.0 and topK\_p = 1 to avoid randomness. Detailed training hyper-parameters and cost can be found in Tab.~\ref{tab:hyper_parameter_training_settings}.

\begin{table*}[h]
\centering
\resizebox{1\textwidth}{!}{
\begin{tabular}{cccccccc}
\toprule
Model Size           & Data Size    & \# GPUs & Epoch & LR & Batch Size & Max Length & Training Time (Minutes) \\ \midrule
7B & 9K & 8 & 3 & 2e-5 & 128 & 512 & 13.33 \\
7B & 52K & 8 & 3 & 2e-5 & 128 & 512 & 161.40 \\ \midrule
13B & 9K & 8 & 5 & 1e-5 & 128 & 512 & 43.16 \\ 
13B & 52K & 8 & 5 & 1e-5 & 128 & 512 & 219.48 \\ \bottomrule
\end{tabular}
}
\caption{Detailed hyper-parameter settings and costs.}
\label{tab:hyper_parameter_training_settings}
\end{table*}

\subsection{Performance of Benchmark on Varying Training Samples}\label{sec:varying_benchmark}
{In this section, we conduct an ablation study to analyze the impact of varying training data proportions on the performance of LLMs across different benchmarks. Instead of ablating components of a model architecture, we systematically evaluate models trained on increasing subsets of the full training data: 3K, 6K, and 9K examples. This allows us to isolate the effect of training data size on the model's ability to generalize to unseen tasks, as measured by MMLU, BBH, and Hellaswag. By comparing these results to the performance of a model trained on the complete dataset (52K), we aim to quantify the performance gains achieved with larger training sets and identify potential saturation points or diminishing returns. This analysis provides insights into the data efficiency of LLMs and the importance of training data scale for achieving optimal benchmark performance.}

{Table~\ref{tab:app_varying} presents a comprehensive performance comparison across several few-shot learning methods: Full (utilizing a larger 52K dataset), Random (subsampling to 3K, 6K, and 9K), Alpagasus (3K, 6K, and 9K), IFD (3K, 6K, and 9K), and our proposed approach, ``Ours'' (evaluated on 3K, 6K, and 9K subsets). The evaluation spans diverse and challenging benchmarks, including MMLU (assessed in both 0-shot and 5-shot settings), BBH, and Hellaswag. The results highlight the performance trends as the size of the training data increases from 3K to 9K, providing insights into the scalability and effectiveness of each method on different tasks. Notably, the final column for each data size (9K) emphasizes the performance of ``OURS'' often demonstrating competitive or superior results in these higher data regimes. The average performance across all tasks is also provided at the bottom, offering a holistic view of each method's overall effectiveness.}

{\begin{table*}[h!]
\centering
\caption{Performance Comparison}
\label{tab:performance}
\resizebox{1\textwidth}{!}{
\begin{tabular}{l c | c c c  c |c c  c c |c  c c c}
\toprule
Benchmarks & Full & Random & Alpagasus & IFD & OURS & Random & Alpagasus & IFD & OURS & Random & Alpagasus & IFD & OURS \\
& (52K) & (3K) & (3K) & (3K) & (3K) & (6K) & (6K) & (6K) & (6K) & (9K) & (9K) & (9K) & (9K) \\
\midrule
MMLU-0-Shot & 22.09 & 21.10 & 22.00 & 21.72 & \textbf{22.30} & 21.80 & 22.40 & 22.10 & \textbf{22.70} & 22.51 & 23.25 & 23.12 & \textbf{23.82} \\
MMLU-5-Shot & 45.45 & 44.00 & 44.70 & 44.80 & \textbf{45.60} & 45.00 & 45.70 & 45.30 & \textbf{46.10} & 46.38 & 46.74 & 46.10 & \textbf{47.44} \\
BBH & \textbf{32.10} & 30.50 & 30.80 & 31.40 & 31.20 & 30.90 & 31.10 & 31.60 & 31.50 & 31.38 & 31.32 & 31.08 & 30.97 \\
Hellaswag & 69.97 & 69.00 & 69.80 & 69.30 & \textbf{70.10} & 69.50 & \textbf{71.20} & 69.65 & 70.50 & 70.99 & \textbf{71.07} & 70.55 & \textbf{71.07} \\
\midrule
Average & 42.15 & 41.15 & 41.83 & 41.53 & \textbf{42.30} & 41.80 & 42.60 & 41.97 & \textbf{42.70} & 42.82 & 43.10 & 42.71 & \textbf{43.33} \\
\bottomrule
\end{tabular}
}\label{tab:app_varying}
\caption{The benchmark results of models fine-tuned on
different subsets selected by corresponding methods.}
\end{table*}}

\begin{table*}[t]
\centering
\resizebox{1\textwidth}{!}{
\begin{tabular}{c|c|ccc|ccc|ccc|ccc}
\toprule
\multirow{2}{*}{Dataset} & \multirow{2}{*}{Selector} & \multicolumn{3}{c}{$L_{B}=10, R=1$} & \multicolumn{3}{c}{$L_{A}=10, R=1$} & \multicolumn{3}{c}{$R=1$} & \multicolumn{3}{c}{$L_{A}=L_{B}=10$} \\
 &  & $L_{A}=20$ Win & $L_{A}=10$ Win & TIE & $L_{B}=10$ Win & $L_{B}=20$ Win & TIE & $L_{A}=L_{B}=10$ Win & $L_{A}=L_{B}=10$ Win & TIE & $R=1$ Win & $R=2$ Win & TIE \\ \midrule \midrule
Sinstruct & \multirow{4}{*}{Qwen2.5-72B-Instruct} & \textbf{102} & 93 & 57 & 89 & \textbf{94} & 69 & 68 & \textbf{116} & 68 & 75 & \textbf{111} & 66 \\
Vicuna &  & \textbf{33} & 23 & 24 & 32 & \textbf{36} & 12 & 26 & \textbf{39} & 15 &  23 & \textbf{40} & 17 \\
Koala &  & \textbf{72} & 56 & 52 & \textbf{69} & {65} & 46 & 49 & \textbf{90} & 41 &  59 & \textbf{81} & 40 \\
Wizardlm &  & \textbf{86} & 81 & 51 & 64 & \textbf{100} & 54 & 76 & \textbf{89} & 53 &  57 & \textbf{98} & 63 \\ \midrule
Sinstruct & \multirow{4}{*}{GPT3.5} & \textbf{105} & 82 & 57 & 88 & \textbf{94} & 70 & 74 & \textbf{122} & 56 & 78  & \textbf{109} & 65 \\
Vicuna &  & \textbf{41} & 27 & 12 & \textbf{32} & \textbf{32} & 16 & 25 & \textbf{47} & 56 & 19 & \textbf{50} & 11 \\
Koala &  & \textbf{73} & 65 & 42 & 63 & \textbf{67} & 50 & 55 & \textbf{94} & 31 &  53 & \textbf{80} & 47 \\
Wizardlm &  & \textbf{96} & 71 & 51 & 68 & \textbf{85} & 65 & 74& \textbf{94} & 50 &  58 & \textbf{101} & 59 \\ \bottomrule
\end{tabular}
}\caption{Ablation study on the influence of the lengths of sets $\mathcal{A}'$ and $\mathcal{B}'$, and the number of selection rounds. The numbers indicate how many times a configuration wins over another in pairwise comparisons.}
\label{app:results_ablation}
\end{table*}

\subsection{Ablation Study}\label{sec:ablation_study}

\paragraph{Ablation Study} Intuitively, a larger $L_A$ and $L_B$ results in $\mathcal{A}'$ and $\mathcal{B}'$ containing more elements, which increases the number of candidates and simultaneously elevates the difficulty of assessing diversity among them. This expansion in the candidate subsets provides the LLM with more options to consider, potentially leading to the selection of samples with greater diversity. Our experimental results are summarized in Tab.~\ref{app:results_ablation}, where we compare different configurations in terms of their ability to select better samples. We report the number of times one configuration wins over another, indicating superiority in diversity and quality of the selected subsets. We conduct evaluations using various LLMs as selectors (Qwen2.5-72B-Instruct and GPT3.5) on multiple datasets (e.g., Sinstruct, Vicuna, Koala, WizardLM). From the results in Tab.~\ref{app:results_ablation}, we observe several key trends:
\begin{enumerate}

    \item \textbf{Effect of the initialization of $\mathcal{A}'$}: {In the preceding experiment, the $\mathcal{A}'$ was initialized through random sampling. It is hypothesized that a more guided initialization, such as employing K-means clustering, could lead to a more advantageous final subset selection and consequently improved model performance. In this section, we experimentally investigate the impact of various initialization strategies for $\mathcal{A}'$ on the characteristics of the ultimately chosen subset. The results presented in Table~\ref{tab:app_ablation_data_char} indicate that utilizing K-means and KNN for the initialization of $\mathcal{A}'$ does indeed result in slightly enhanced Diversity Measurements, as evidenced by metrics like TTR and MTLD. However, when considering Quality Measurement, the Random Initialization approach achieves comparable performance to both K-means and KNN, even outperforming them in terms of Helpful Score. These findings suggest that our method exhibits a degree of robustness with respect to different initializations of $\mathcal{A}'$. We posit that the underlying reason for this resilience stems from our experimental constraint of setting the size of $\mathcal{A}'$ to a maximum of 20, as a consequence, the initial selection within a relatively small pool might not drastically alter the final refined subset obtained through our subsequent selection process.}

    \item \textbf{Effect of Length of Set $\mathcal{A}'$}: Increasing the length of the selected subset $\mathcal{A}'$ from 10 to 20 generally leads to better performance. For instance, on the Sinstruct dataset using Qwen2.5-72B-Instruct as the selector, $\mathcal{A}'$ of length 20 wins 102 times versus 93 times for length 10, with 57 ties. This suggests that providing more context from the current selected set helps the LLM make more informed decisions, enhancing diversity.
    \item \textbf{Effect of Length of Candidate Set $\mathcal{B}'$}: Similarly, a larger candidate set $\mathcal{B}'$ (length 20) improves performance compared to a smaller candidate set (length 10). For example, on the WizardLM dataset with GPT3.5 as the selector, $\mathcal{B}'$ of length 20 wins 85 times versus 68 wins for length 10. A larger candidate pool offers more options for the LLM to select diverse and high-quality samples.
    \item \textbf{Combined Effect of Lengths of $\mathcal{A}'$ and $\mathcal{B}'$}: When both $\mathcal{A}'$ and $\mathcal{B}'$ are increased to length 20, we observe a cumulative positive effect. The configuration with both sets at length 20 often wins more comparisons against the configuration where both are at length 10. This indicates that simultaneously increasing both sets' lengths amplifies the benefits in diversity and quality.
    \item \textbf{Multi-Round Selection Strategy}: Implementing multiple rounds of selection shows a consistent advantage. The ``Round 2'' configuration, where a second round of selection is performed on an initially larger subset, often outperforms the ``Round 1'' configuration. For instance, on the Sinstruct dataset with Qwen2.5-72B-Instruct, Round 2 wins 111 times versus 75 wins for Round 1. This demonstrates that iterative refinement through multiple selection rounds effectively enhances the final subset's diversity and quality.
    \item \textbf{Consistency Across Models and Datasets}: These trends are observed across different LLM selectors (Qwen2.5-72B-Instruct and GPT3.5) and datasets (Sinstruct, WizardLM, Vicuna, Koala). This consistency suggests that the benefits of larger set sizes and multi-round selection are generalizable. Besides, our method is not limited to specific models or datasets.
\end{enumerate}

{\begin{table}[h]
\centering
\resizebox{0.8\textwidth}{!}{
\begin{tabular}{c|ccc|cc|c}
\toprule
\multirow{2}{*}{Method} & \multicolumn{3}{c|}{Diversity}  & \multicolumn{2}{c|}{Quality} & \multirow{2}{*}{\#Tokens} \\
& TTR $\uparrow$ & MTLD $\uparrow$  & SDI $\downarrow$ & DEITA $\uparrow$ & Helpful $\uparrow$  &  \\ \midrule \midrule
Full (52K) & 95.47 & 7.9352  & 0.1067 & 2.765 & 1.314 & 11.35\\
Random & 95.46 & 7.9421  & 0.1068 & 2.764 & 1.397 & 11.33\\
AlpaGasus & 96.07 & 8.0508 & 0.1086& 2.969 & 2.025 & 10.93 \\
IFD & 96.05 & 8.0444  & 0.1091 &  3.127 & 2.456 & 10.80\\
OURS + Random & {96.24} & {8.4449} & \textbf{0.1035} & {3.210} & \textbf{2.703} & 11.29 \\ 
OURS + Kmeans & \textbf{96.58} & \textbf{9.0125} & {0.1057} & {3.092} & {2.613} & 11.57 \\
OURS + KNN & {96.39} & {8.8721} & {0.1048} & \textbf{3.278} & {2.581} & 11.98 \\ \bottomrule
\end{tabular}
}
\caption{Performance of different methods on various text diversity and quality metrics. The OURS method shows the best performance across multiple metrics.}
\label{tab:app_ablation_data_char}
\end{table}}

In summary, the ablation study confirms that increasing the lengths of sets $\mathcal{A}'$ and $\mathcal{B}'$ allows the LLM to consider more context and a wider range of candidates, leading to the selection of samples with greater diversity and quality. Additionally, employing a multi-round selection strategy further refines the subset by allowing the LLM to iteratively focus on the most promising candidates. However, it is important to balance these benefits with computational considerations, as larger sets and additional rounds may increase inference time. Selecting appropriate lengths for $\mathcal{A}'$ and $\mathcal{B}'$, as well as an optimal number of selection rounds, is crucial for maximizing performance while maintaining efficiency.


\subsection{Explanation To Characteristics Metrics}\label{sec:characteristics_metrics}
\paragraph{Type-Token Ratio} The Type-Token Ratio~\cite{article} (TTR) represents the relationship between the quantity of distinct words (types) emerging in a text and their occurrence frequencies. The count of unique words within a text is conventionally termed the number of types. It's worth noting that some of these types recur. The value of the TTR spans from 0 to 100. A larger number of types relative to the total number of tokens (resulting in a higher ratio value) indicates a richer vocabulary. In other words, the text exhibits greater lexical diversity.
The Type-Token Ratio is computed as follows:
\begin{equation}
    \text{TTR} = \frac{\text{Number of unique types}}{\text{Number of tokens}} * 100
\end{equation}
\paragraph{Measure of Textual Lexical Diversity} The Measure of Textual Lexical Diversity~\cite{mccarthy2010mtld} (MTLD) is a metric designed to assess the lexical diversity of a text while mitigating the influence of text length, a limitation often encountered with traditional measures like the Type-Token Ratio (TTR). Unlike TTR, which can vary significantly with the length of the text, MTLD remains relatively stable, providing a more consistent evaluation of lexical diversity across texts of varying lengths. The MTLD value is determined by dividing the total number of words by the total number of factors, effectively representing the average length of word strings that maintain the desired TTR threshold. The MTLD is computed as follows:
\begin{equation}
    \text{MTLD} = \frac{\text{Total number of words in the text}}{\text{Number of factors}}
\end{equation}
A higher MTLD value indicates greater lexical diversity, as it implies longer sequences of words are required before the TTR falls below the threshold, reflecting richer and more varied vocabulary. By accounting for both the range and distribution of vocabulary, MTLD provides a robust assessment of lexical diversity that is less sensitive to text length compared to other metrics.

\paragraph{Simpson's Diversity Index} The Simpson Diversity Index~\cite{1997Measurement} (SDI), originally developed in ecology to measure biodiversity, can be effectively applied to textual analysis to assess lexical diversity. In textual diversity analysis, unique words are treated as species, and their frequencies as species abundances. The SDI quantifies the probability that two randomly selected words from a text are different and is calculated using the formula:
\begin{equation}
    D = \sum_{i=1}^{N} p_i^2,
\end{equation}
where $p_i$ represents the proportion of the $i$-th word type in the text, calculated as $p_i = \frac{n_i}{N_t}$. $n_i$ is the frequency of the $i$-th word and $N_t$ the total number of words. Values close to 1 indicate higher homogeneity, thus lower diversity and values close to 0 indicate higher variability, thus higher diversity. This measure is particularly useful for understanding the complexity and style of a text, as it accounts for both the number of unique words and their frequency distribution.

\paragraph{Helpful Reward Score} Reward scores usually help the model learn to maximize helpfulness by adjusting its parameters based on these scores, ensuring that the training data selected is optimally beneficial for improving performance. We adopt a reward model that is trained on the Anthropic Helpful Harmless dataset and achieves a test accuracy of over 75\%~\cite{dong2023raftrewardrankedfinetuning}, where a higher score indicates the better response quality regarding to its input instruction.

\subsection{In-Depth Analysis of Efficiency}\label{app:in-deep_analysis}

\subparagraph{Time Complexity (TC) and Number of Iterations (NoI)} SOTA methods that rely on LLMs for selection (e.g., IFD, Alpagasus, and DeITA) typically employ pointwise scoring approaches. These methods have a time complexity of $\mathcal{O}(N \times T_{\text{LLM}})$, where $N$ denotes the total number of samples in the full training dataset and $T_{\text{LLM}}$ represents the time required for the LLM to perform a single inference. This implies that to select a subset of size $K$, these methods need to process all $N$ samples, resulting in $N$ data accesses. In contrast, as discussed in Section~\ref{sec:discussion}, our method operates with a time complexity of $\mathcal{O}(K \times T_{\text{LLM}})$. This means we perform selections proportional only to the desired subset size $ K $, embodying the principle of ``selecting only what is necessary''. When $K \ll N$, our method demonstrates significantly enhanced efficiency compared to pointwise scoring methods. This efficiency gain largely aligns with the goal of sample selection: to train effective models using less data and computational resources.

\subparagraph{Practical Time Consumption (PTC)} Beyond theoretical analysis, we provide empirical comparisons of the actual time consumed during the selection process. Based on the Alpaca dataset and under identical hardware conditions, we reproduce IFD and Alpagasus for fair comparison. For both IFD and our method, we utilize Qwen2.5-7B-Instruct and set $ L_{A} = L_{B} = 20 $ in our method. For Alpagasus, we use the official OpenAI API to query GPT3.5. To conserve time and computational resources, we measure the time required to perform selection (for our method) or scoring (for IFD and Alpagasus) on 1,000 samples. We report the total time taken for the process. 

Additionally, we also explore how varying the value of $L$ in our method affects the overall practical time consumption (PTC), providing insights into the trade-offs between selection granularity and computational efficiency. Specifically, under the same hardware environment, we record the time consumption for 100 selections under different values of $L_A$ and $L_B$, when employing Llama3.1-8B-Instruct as the selector. The results in Tab. \ref{tab:time_increasing} show that our method does not experience a sharp increase in time consumption as $L_A$ and $L_B$ rise. Although the practical time consumption generally grows with larger $L_A$ and $L_B$ values, the rate of increase is relatively moderate. This indicates that within a certain range, we can adjust $L_A$ and $L_B$ to balance the selection and computational cost without being overly burdened by excessive time expenditure.

\begin{table}[h]
\centering
\resizebox{0.6\textwidth}{!}{
\begin{tabular}{cccccc}
\toprule
$L_A = L_B = $ & 5 & 10 & 15 & 20 & 30 \\ \midrule\midrule
PTC (Second) & 25.42 & 40.84 & 69.95 & 103.62 & 108.91 \\ \bottomrule    
\end{tabular}
}
\caption{Detailed hyper-parameter settings and costs.}
\label{tab:time_increasing}
\end{table}

In summary, results of these metrics prove that our method strikes an acceptable balance between enhanced performance and computational costs. The results indicate that our approach is more efficient, particularly when the desired subset size $ K $ is much smaller than the full dataset size $ N $, fulfilling the objective of achieving better models with less data and reduced time investment.

\subsection{Scaling-up with Larger Dataset}\label{sec:medical_exp}

\paragraph{Datasets} We adopt HuatuoGPT-sft-data-v1\footnote{\scriptsize\url{https://huggingface.co/datasets/FreedomIntelligence/HuatuoGPT-sft-data-v1}} as SFT samples that includes 226K Chinese medical QA pairs as SFT samples and evaluate the performance on Chinese Medical Benchmark (CMB)~\cite{wang2024cmbcomprehensivemedicalbenchmark} that is a comprehensive and all-encompassing Chinese medical quiz assessment benchmark, which contains 12K human-annotated five-option multi-choice questions and cover six aspects in benchmarking a medical LLM. The Tab.~\ref{tab:hyper_parameter_training_settings_medical} highlights the training hyper-parameters.

\begin{table*}[!h]
\centering
\resizebox{0.6\textwidth}{!}{
\begin{tabular}{cccccc}
\toprule
Model Size & \# GPUs & Epoch & LR & Batch Size & Max Length \\ \midrule \midrule
7B  & 2*8 & 3 & 2.5e-5 & 128  & 2048 \\ \bottomrule
\end{tabular}
}
\caption{Detailed hyper-parameter settings and costs.}
\label{tab:hyper_parameter_training_settings_medical}
\end{table*}

\section{Case Analysis}\label{case_ana}

\begin{figure}[h]
    \centering
    \includegraphics[width=1\linewidth]{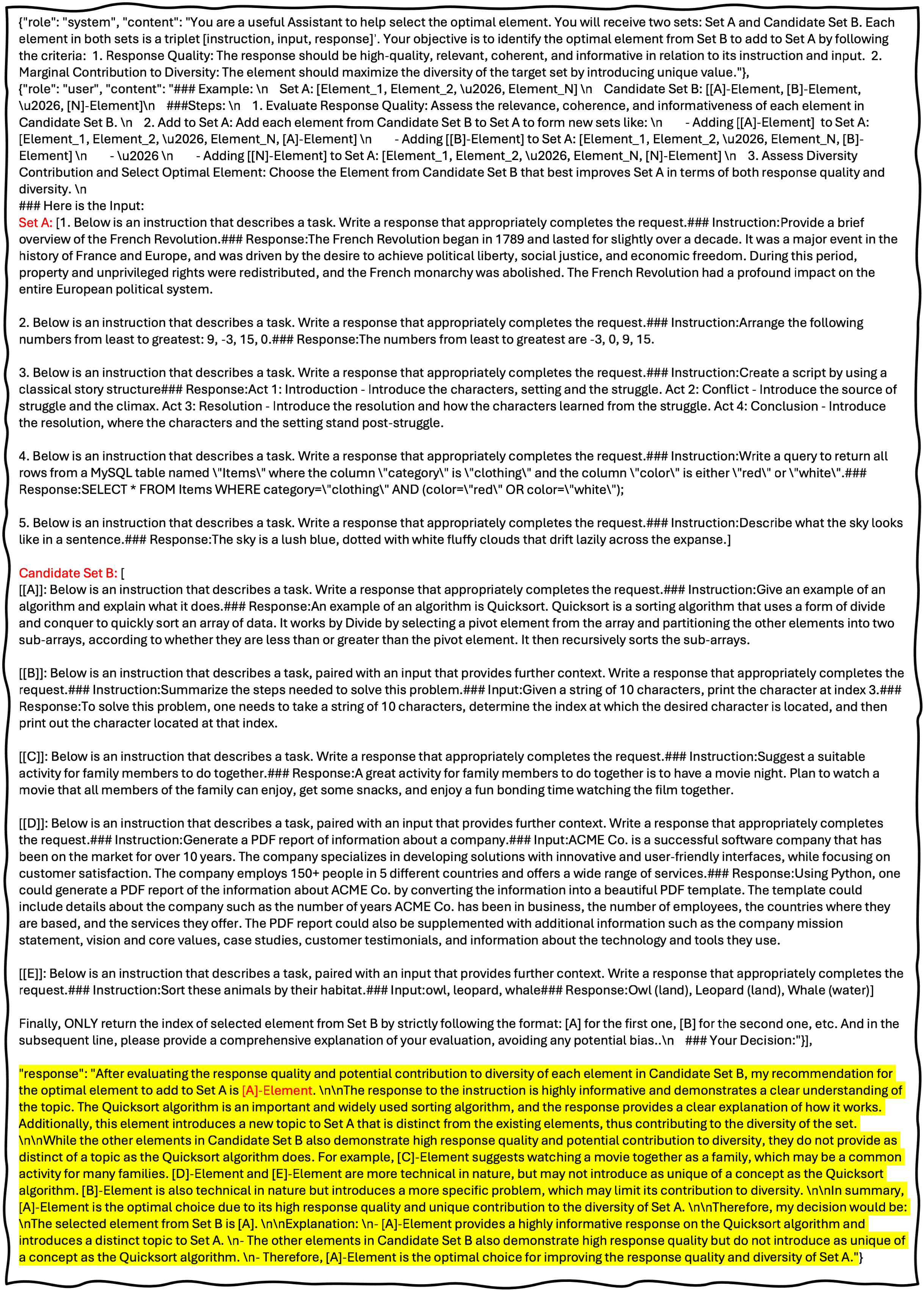}
    \caption{Selection case analysis.}
    \label{fig:case_1}
\end{figure}

\begin{figure}[h]
    \centering
    \includegraphics[width=1\linewidth]{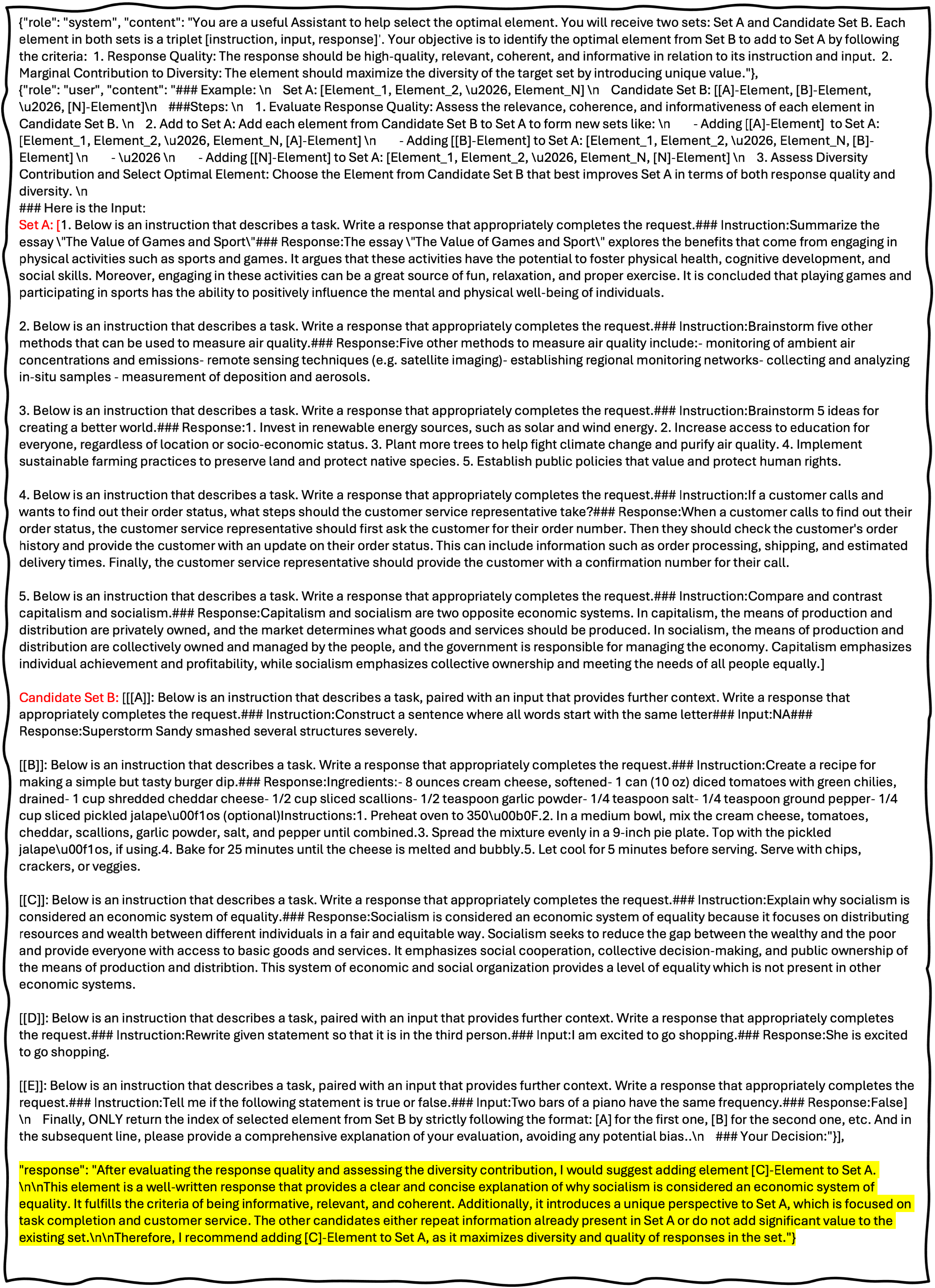}
    \caption{Selection case analysis.}
    \label{fig:case_2}
\end{figure}

\begin{figure}[h]
    \centering
    \includegraphics[width=1\linewidth]{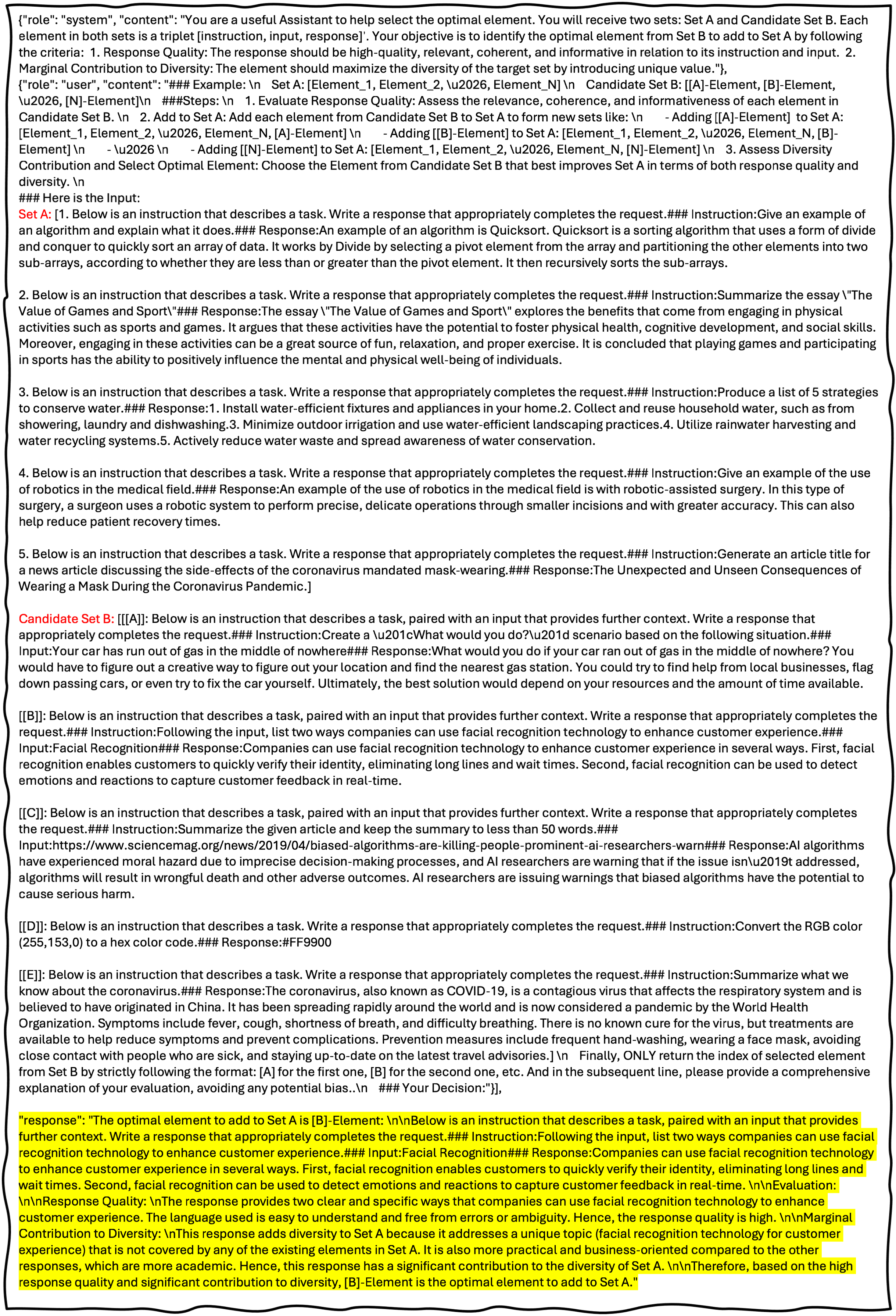}
    \caption{Selection case analysis.}
    \label{fig:case_3}
\end{figure}

In this section, we provide three case selection analyses when employing GPT3.5 as the selector. As shown in Fig.~\ref{fig:case_1}, Fig.~\ref{fig:case_2}, and Fig.~\ref{fig:case_3}, we find LLMs can effectively make a reasonable selection choice given the selection prompt that includes the selected set $A$ and the candidate set $B$. For example, as shown in Fig.~\ref{fig:case_1}, the selector recommends \textit{[A]-Element} for the optimal element to add to the set $A$, by providing a highly informative response on the Quicksort algorithm and introducing a distinct topic to Set $A$. The other elements in Candidate Set $B$ also demonstrate high response quality but do not introduce as unique of a concept as the Quicksort algorithm in [A]-Element.

\end{document}